# Identifying Human Edited Images using a CNN

Jordan Lee[1], Willy Lin[2], Konstantinos Ntalis[2], Anirudh Shah[2], William Tung[2], and Maxwell Wulff[2]

[1]Department of Computer Science, Cornell Tech
[2]Department of Operations Research & Information Engineering, Cornell Tech

## Abstract

Most non-professional photo manipulations are not made using propriety software like Adobe Photoshop, which is expensive and complicated to use for the average consumer selfie-taker or meme-maker. Instead, these individuals opt for user friendly mobile applications like FaceTune and Pixlr to make human face edits and alterations. Unfortunately, there is no existing dataset to train a model to classify these type of manipulations. In this paper, we present a generative model that approximates the distribution of human face edits and a method for detecting Facetune and Pixlr manipulations to human faces.

## 1 Introduction

In the current social media dominated times, there are enormous amounts of content and media being generated for social media consumption. This has led to concerns over the rise of manipulated content such as 'deepfakes' and 'fake news' and the dangers associated [1, 2]. However, these types of content often contain artifacts that are relatively easily identifiable to the human eye, and so in this report we will focus on more subtle image edits that are not perceivable to the human eye but can be identified accurately by a machine learning or deep learning algorithm [3]. These include those done using Adobe Photoshop and Facetune and whilst such editing techniques have allowed for creative exploration, they can also be associated with significant negative implications including body image issues and the spreading of misinformation [4]. We will focus our work on these types of edits. The majority of non-professional photo edits are not made using Adobe Photoshop but made using cheaper and easier-to-use apps such as Facetune and Pixlr. In this work, we generated a dataset of edited faces using a generative model, and then developed a classifier to determine if an input image was edited or not, using our generated images and a dataset of unedited faces as training data. Our investigation followed the framework of Wang et al. research, however we sought to develop a model with increased generalizability compared to their model and also explored other methods of data collection/generation [4]. All of the code associated with this project can be found on GitHub [1].

## 2 Related Work

### 2.1 Identifying Human Edited Images

There has been limited published work in this area as most research has been focused on machine generated edited images and videos. Wang et al. developed a binary classifier that was able to detect edited images with a 90% accuracy, however upon testing the model we were not able to replicate this level of accuracy with our dataset. It seems that this classifier did not generalize well as it was trained on Adobe Photoshop face aware liquify tool edited images and did not take into account other types of edits or other editing apps. In their work, Wang et al. also developed a model to identify the location of the edit on the image [4].

---

[1] https://github.com/JordanMLee/deeplearningproject

Manu et al. focused on spliced images and forgeries using classification of texture patterns on images and image quality artifacts generated during the tampering operation along with a support vector machine classifier to detect altered images [5]. Further work in this area involves taking a more targeted approach to feature extraction by using bilateral filtering followed by Gray Level Co-occurrence Matrices [6].

Zhou et al. proposed a two stream neural network technique in which one stream detects lower level irregularities between image patches and the other stream detects altered face images [7].

### 2.2  Synthesizing Images using Generative Models

Over the last few years there has been a significant amount of research in the area image generation using generative models. This is especially the case as annotated image data for computer vision tasks is hard to come by, and so researchers have focused on methods of augmenting datasets using generated images [8].

Zhou et al. propose a method to generate images using a GAN with relevance feedback which allows a user to give feedback to the model allowing the user to retrieve the generated images that are most similar to the target image [9]. He at al. hypothesized that the encoder-decoder architecture utilized in GANs means the editing of facial attributes is attained by decoding the latent representation of the source face given the wanted facial features, and that this can lead to loss of information. They suggested another technique - applying an attribute classification constraint to the generated image, rather than the typically applied attribute-independent constraint on the latent representation, and so only making user desired changes to the facial features. Furthermore, they proposed that the image reconstruction maintains attribute-excluding details and achieved realistic to the human-eye edits, leading to the attGAN [10].

Another optimization in this field is the use of triple translation GAN (TTGAN) with a multi-layer sparse representation to improve on existing issues with GAN generated images including low quality facial features, unstable model optimization and changing of face identity [11]. Hsu et al. developed an improved implementation of Disentangled Representation-learning GAN by using Wasserstein loss in training the discriminator and they were able to achieve improved quality in synthesized facial images[12]. Bolluyt and Comaniciu introduced a new method of generating images containing multiple target classes by learning the interrelationships of the different desired classes, which could produce realistic synthesized images. They proposed a new Conditional Deep Convolutional GAN architecture that prevented mode collapse and self-corrected during training to prevent suboptimal results [13].

## 3  Methodology

### 3.1  Efficiency of the original model

We started by studying how well the method proposed in [4] works for warping applied to human face images using tools other that Photoshop. Following the editing methods discussed later in this section and also in Section 4 - Data Collection, we discovered that the binary classifier in [4] performs poorly in this task. Specifically, for a validation set of 500 manually edited images, the classifier correctly predicted whether the image was edited or not with 60% accuracy. Such an example is provided Figure 1. By performing this experiment, we realized that the trained model is over-fitting the training data distribution and can only be reliable for professional edited photos using Photoshop.

### 3.2  Augmenting the training set

We therefore attempted to use the framework proposed in [4] but develop a more diverse training set in order to be able to capture not only more types of edits but also, what we believe to be more common distributions of edited images, generated by tools other than Photoshop. Given the lack of existing datasets for this task, we decided to develop our own. We started by manually editing face images from different public datasets shown in Table 1. Then, using three different tools, Adobe Lightroom, Facetune and Pixelmator, we performed several face edits as shown in Table 1. As it would be impossible to manually generate enough to robustly train a classifier, we attempted to train a GAN in order to generate the classifier training set. Using non edited and the corresponding manually edited images, we generated 149230 fake images using a pre-trained GAN (DCGAN) [14]. These generated images are either fake unedited or fake edited. In this way, we are asserting that we can approximate the distribution of edited images and thereby more efficiently train a binary classifier to detect more general edits. All of our GAN generated images passed the Haar Cascade face classifier algorithm and so we decided to proceed with using these images as our training data [15].

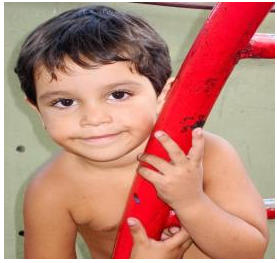 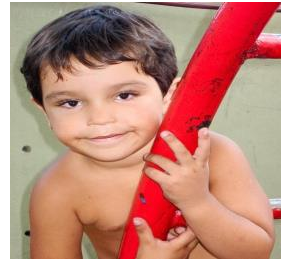

(a) Original image　　　　　　　　　　　　　　　　　　　　(b) Edited image

Figure 1: An example where the classifier developed in [4] fails to determine the edited image.

### 3.3    Real-or-edited classification

Having built the dataset with fake reals and fake edited images, we proceed with training the binary classifier. The original paper uses a Dilated Residual Network variant (DRN-C-26) [16] that was originally designed for segmentation analysis [4]. We used this as a baseline and experimented with other architectures such as ResNeXt [17] that improved our results. For both architectures we used pre-trained networks trained on similar tasks with only additional linear layers being trainable when needed.

### 3.4    Experimentation

Except for fine tuning the networks and experimenting for the most appropriate architecture, we further experimented with the composition of the dataset. One important decision is resolution. As the authors of [4] note, high-resolution models enable preservation of low-level details, potentially useful for identifying fakes. On the other hand, a lower-resolution model potentially contains sufficient details to identify fakes and can be trained more efficiently. As our pipeline involved GAN generated images, the lower resolution option inevitably comes with less realistic fake images.

Finally, we examined the option of including real unedited and real edited images in the training set of the final classifier too. Although promising we didn't move forward with this choice as it is unclear what the binary classifier will eventually learn - distinguishing edited from unedited or real from fake images.

## 4    Data Collection

In order to properly train a model that had a wide use case, we collected unaltered images from a multitude of sources, and altered them through a variety of means. We began with the 500 Edited Validation Images from Wang et al. These were collected from the Flickr Faces HQ (FFHQ) Dataset and altered with Adobe software.[18] 103 images were collected from FFHQ and altered using Adobe Lightroom. These edits focused on smoothing features and enhancing lighting. 960 faces were collected from Real and Fake Face Detection dataset (RFFD). [19] These images are faces with some features swapped out for those features in different images. 218 images were taken from the Helen dataset and altered using Facetune. [20] Facetune is one of the most popular 'selfie' editor applications. [21] These edits focused on feature smoothing, teeth whitening, and head reshaping. 100 faces were taken from the Helen Dataset and edited using Pixelmator. Pixelmator is a popular photo editing application. These edits focused on warping, bumping and/or pinching parts of the original image. Finally, 100 images were scraped off of Facebook, and edited using Facetune. In summary we gathered 2521 images from a variety of sources and altered them by varying means to get a more realistic sense of the scope of human edited images that are employed for social media.

| Number | Source | Editing Software | Edits Performed |
|---|---|---|---|
| 500 | FFHQ | Adobe | Reshape, Smoothing |
| 103 | FFHQ | Adobe Lightroom | Smoothing, Lighting Alteration |
| 960 | RFFD | NA | Feature Swapping |
| 218 | Helen | Facetune | Reshape, Smoothing, Teeth Whitening |
| 100 | Helen | Pixelmator | Warping, Bumping, Pinching |
| 100 | Facebook | Facetune | Reshape, Smoothing, Teeth Whitening |

Table 1: Edited Face Sources

## 5 Model

### 5.1 Adapted DCGAN for Image Synthesis

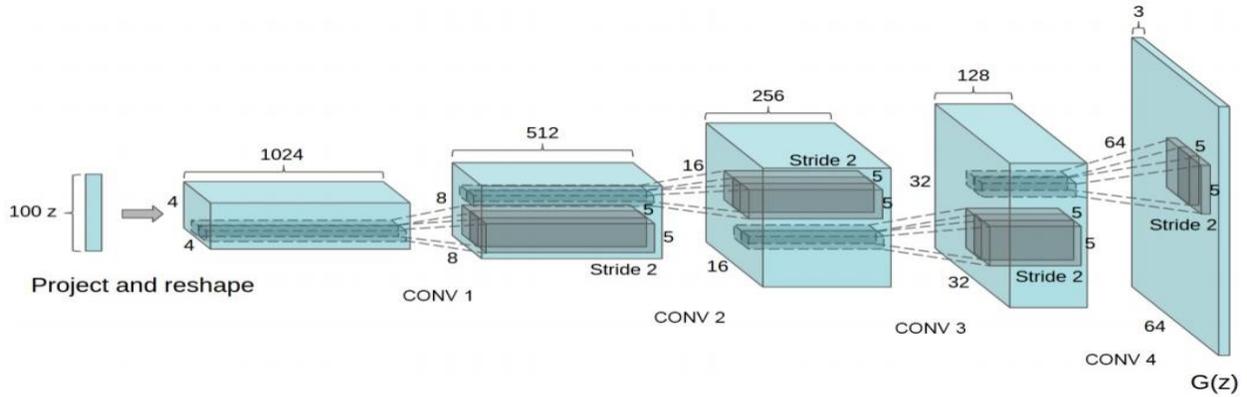

Figure 2: Architecture of generator from DCGAN [14]

The model used for image generation was an adaptation of the DCGAN as proposed by Radford et al.[14], the architecture of the original DCGAN discriminator is shown in Figure 2. This consists of two CNNs with opposing optimization methods, however, in general the convolutional layers aggregate the features of the input image sequentially while reducing spatial dimensionality. Our adaptation to DCGAN was so that it outputted both 64×64 and 128×128 3-channel RGB images.

The adapted DCGAN discriminator for the 64×64 images were as follows: update the feature maps in the first convolutional layer to 64, increase the feature maps by a factor of 2 for every subsequent convolutional layer and use 5 hidden layers. The adapted DCGAN discriminator for the 128×128 images were as follows: update the feature maps in the first convolutional layer to 16 to motivate convergence, increase the feature maps by a factor of 2 for every subsequent convolutional layer and use 6 hidden layers. The following parameters are common to both types of outputs: each hidden layer had a 4×4 convolutional keren with stride length 2, each hidden layer was followed by a batch normalization layer to constrain co-variate shift between mini batches and a Leaky ReLU activation layer, the learning rate was 0.0002.

The adapted DCGAN generator architecture is as follows: 64 feature maps for the first transposeconvolutional layer, and upsample the image by a factor of 2 for every subsequent convolutional layer. In total there were 5 hidden layers transpose-convolutional layers for the 64×64 model and 6 such layers for the 128×128 model, each transpose-convolutional layer was followed by batch normalization layer and a Leaky ReLU activation layer. The input to the generator was $z$-dimensional noise with the latent vector $z$ serving as an arbitrarily low-dimensional starting point to generate new images. The learning rate was set as 0.0002. Both the Generator and Discriminator were optimized using Adaptive Moment Estimation (Adam).

## 5.2 ResNeXt for Classification

Our binary classifier utilized the ResNeXt50 classifier architecture which consists of 43 convolutional layers and was pretrained on the ImageNet-1k dataset [22, 23]. Our classifier utilized the pretrained weights and finetuned the model to our data by replacing the final fully connected layer with a binary classification layer to determine whether an image is edited or not. This model was trained with a learning rate of 0.01 and was also optimized using Adam.

# 6 Results

## 6.1 Training framework

We trained DCGAN and ResNeXt on one Amazon p3.8xlarge EC2 instance which had 4 GPUs. The DCGAN was trained on a dataset consisting of 2521 hand-edited and the classifier on the 14923 generated images. On Figure 3, we display the images that go into the DCGAN training and on Figure 5 the generated ones that are used for the classifier.

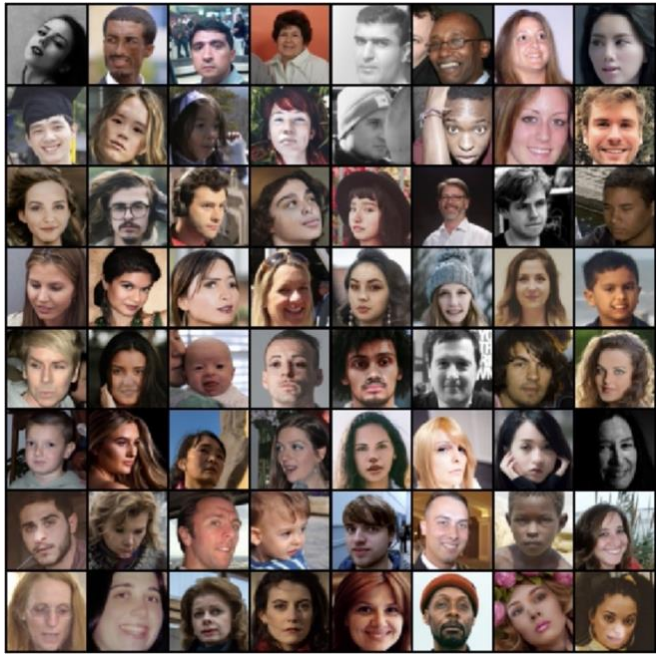

Figure 3: GAN hand-edited training images

## 6.2 GAN training with early stopping

We trained the GAN for 700 iterations on the training set to achieve a generator loss of $< 1.0$. Figure
4 below shows the training loss as a function of epochs. We consistently conducted checkpoints of the model parameters to determine the optimal model weights for generating new training images, given when the generator loss reached $< 1.0$.

We also used the GAN to generate unedited examples by training on the popular celebsA dataset. We trained the 200,000 image dataset for 25 epochs. Training concluded once we visually verified that the GAN generated unedited faces matched the quality of the GAN generated edited faces.

After creating these images, we produced a GAN generated image set containing 149,230 images as shown in Figure 5. This dataset is fully comprised of generated edited images and generated unedited images with random horizontal mirrors. We normalized the input dataset so that the three input channels had values between 0 and 1.

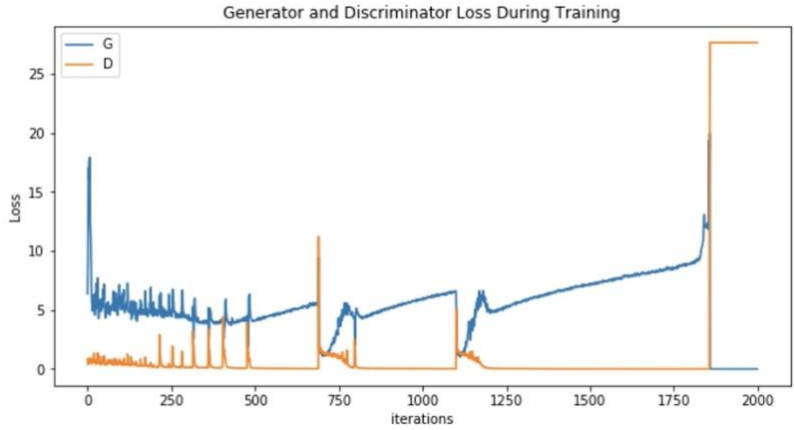

Figure 4: GAN training curves (with early stopping at 700 epochs)

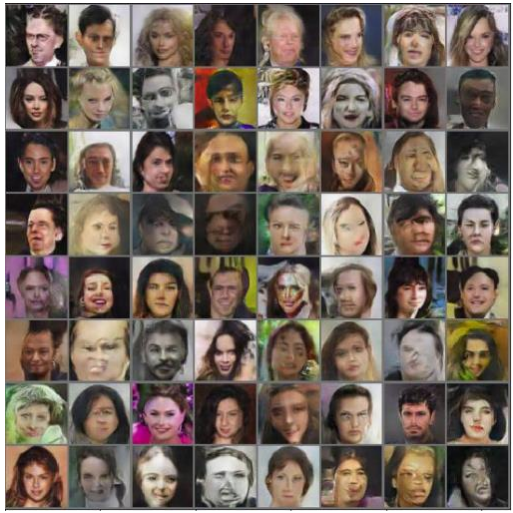

Figure 5: Example Images from Classifier Data set

### 6.3 Classifier training

Finally, we trained the classifier for 50 epochs with a batch-size of 64. The validation loss of the binary classifier is shown in Figure 6. We reserved a subset of our hand-edited images to test the binary classifier and compare it the FALdetector global classifier from Wang et al [4]. The dataset consisted of 333 hand-edited images and 333 unedited images. The main experiments are summarized in Table 2.

| Binary classifier | Accuracy | Precision |
|---|---|---|
| GAN dataset | | |
| ResNeXt | 0.6188 | 0.7653 |
| ResNeXt & real images | 0.5871 | 0.7032 |
| DRN-C-26 | 0.5972 | 0.7365 |
| Dataset from [4] | | |
| ResNeXt | 0.5121 | 0.5347 |
| DRN-C-26 | 0.5045 | 0.5011 |

Table 2: Experimenting with different models and datasets for the binary classification problem.

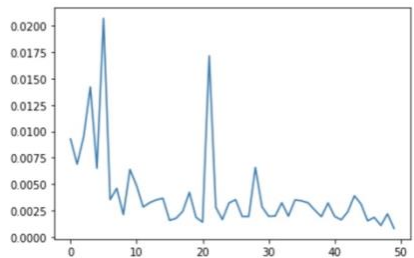

Figure 6: Validation Loss of binary classifier

Note that except from the obvious experiments with different models and datasets, we also tried mixing some of the real photoshopped edited images from the development set of [4], without obtaining a significant increase in accuracy.

## 7 Discussion

Here we outline and discuss some of the challenges encountered throughout the course of our work. These challenges pertain to data collection and training the GAN/Classifiers models.

### 7.1 Data Challenges

When building off the original work of Wang et al. [4], we had difficulty producing an improved classifier without a robust training set. Unfortunately, our main source, Wang et al. [4], does not include a training-set in their project GitHub repository. After conducting a thorough search, we were unable to find a dataset that contained a sufficient number of human-edited faces. The lack of edited images also seemed to trouble Wang et al. [4], as they commissioned photo-shopped edited images. This dilemma of lacking edited image data actually spurred a major directional change in our work with the goal to generate our own images to mimic human manually edited images through common platforms such as FaceTune and Pixlr. To this end, a smaller set of manually edited images was fed into the GAN model in order to mass produce a data-set to be used for the classification task. We chose the method of feeding manual edits into a GAN model because other methods to mass produce an edited dataset, such as scripting Adobe Photoshop, are not representative of how user edited face images. The generative approach seeks to approximate the distribution of commonly made edits to images of the human face using tools like FaceTune and Pixlr, which are two of the most widely-used image editing toolkit application. The FaceTune app is shown below in Figure 7. However, the quality of some of the generative images is fairly low. This is due to the variation in face orientations in the training set. In general, the Generator learns the where to place landmarks like eyes and mouth but has difficulty drawing the contours of the human face. Most GAN datasets are preprocessed so that the orientation of the objects is uniform. This produces good results at the cost of variety.

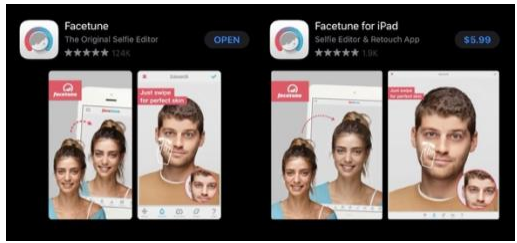

Figure 7: FaceTune

## 7.2 GAN Training

Implementing a GAN model to generate 'fake' edited images presented several challenges on its own.

Firstly, considerable time and computational resources are required to train the GAN model. Such an approach required leveraging several GPU instances through services such as Amazon Web Services (AWS). Even then, several initial AWS setups proved to be too slow or computationally lacking to successfully train models. Aside from computational training limitations, homing in on an optimal GAN model proved challenging. In many instances the GAN model experience mode collapse and would converge on random best fit pixels, common after training for thousands of epochs. Other issues included a lack of convergence of the GAN model. These training issues are highlighted in Figure 8 and Figure 9. We attempted to rectify these issues using some well-known GAN "tricks" like unrolling the GAN and using a different loss function. One specific method we tried was Wasserstein Loss with Gradient Clipping, however the quality of the images was about the same as the DCGAN with Binary Cross Entropy Loss. Obviously, we hypothesize that with better quality generated images our classifier would also thus be better able to discern real non-edited images from fake edited images (GAN produced). This again circled back to our initial issue of creating a robust classifier without a solid dataset.

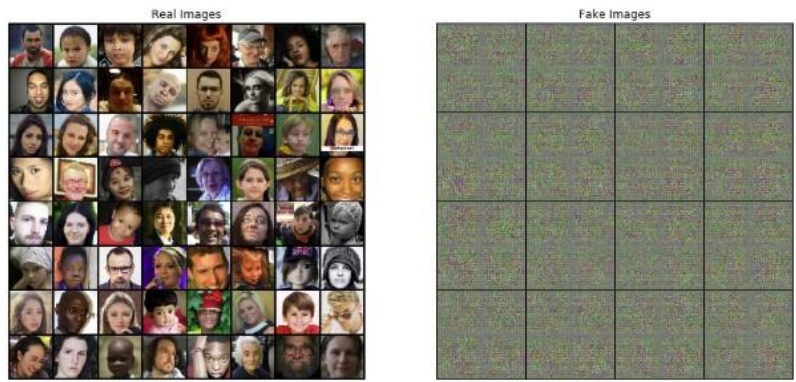

Figure 8: Generator collapse after 2400 epochs

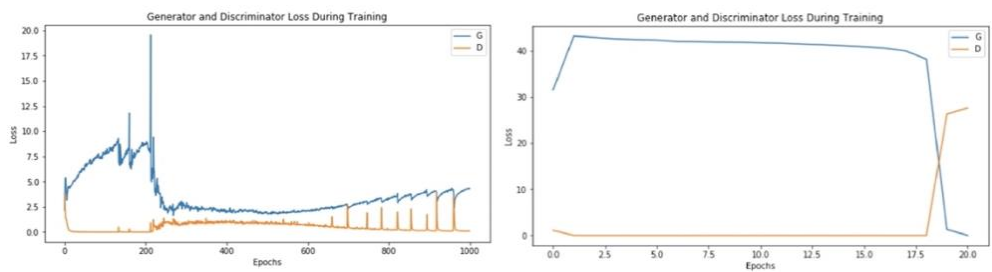

Figure 9: GAN non-convergence and overfit examples

## 8  Conclusion

By exploring the works of Wang et al. [4], we were able to better generalize the detection of manipulations from more mainstream and commonly used software than the originally explored Adobe Photoshop. In doing so, we can hopefully further mitigate the negative effects on body image associated with altered photos. We can seek to continue generalizing our classifier to accurately predict alterations with other software by increasing the training set to incorporate images from many different photo altering packages.

Additionally, by supplementing our dataset with 149230 GAN-generated edited images, we were able to introduce a larger number of training examples which ultimately improved our accuracy and precision. This method of producing supplemental images produced relatively adequate results, which further suggests that GAN models may be used to improve training sets to improve the efficacy of classifiers or other models. With more pre-GAN training data, we may be able to improve the relative quality of the GAN generated altered images, and thus the generic classifier.